\documentclass{article}

     \usepackage[preprint,nonatbib]{neurips_2020}

\usepackage[utf8]{inputenc} 
\usepackage[T1]{fontenc}    
\usepackage{hyperref}       
\usepackage{url}            
\usepackage{booktabs}       
\usepackage{amsfonts}       
\usepackage{nicefrac}       
\usepackage{microtype}      
\usepackage{todonotes} 
\usepackage{subcaption}

\usepackage[]{acronym}

\usepackage{algorithm}
\usepackage{algorithmicx}
\usepackage{amsmath}
\usepackage{algpseudocode}

\usepackage{tabularx,ragged2e}
\usepackage{comment}
\usepackage{multirow}
\usepackage{float}

\newcommand{\etal}{\textit{et al.}}
\newcommand{\eg}{\textit{e.g.}}
\newcommand{\ie}{\textit{i.e.}}

\DeclareMathOperator{\X}{\mathcal{X}}
\DeclareMathOperator{\Y}{\mathcal{Y}}
\DeclareMathOperator{\Loss}{\mathcal{L}}

\acrodef{AI}{Artificial Intelligence}
\acrodef{AUC}{Area Under the Curve}
\acrodef{ROC}{Receiver Operating Characteristic}

\acrodef{PP}{Predicted Positives}
\acrodef{FP}{False Positive}
\acrodef{TP}{True Positive}
\acrodef{FPR}{False Positive Rate}
\acrodef{TPR}{True Positive Rate}
\acrodef{BCE}{Binary Cross Entropy}
\acrodef{GDPR}{General Data Protection Regulation}
\acrodef{JOEL}{Jointly learned cOncept-based ExpLanations}

\acrodef{ML}{Machine Learning}
\acrodef{NN}{neural network}
\acrodef{FFNN}{feed foward neural network}
\acrodef{DNN}{deep neural network}
\acrodef{DL}{deep learning}

\acrodef{XAI}{Explainable Artificial Intelligence}

\title{Teaching the Machine to Explain Itself\\ using Domain Knowledge}

\author{%
  Vladimir Balayan \\
  Feedzai \\
  \texttt{vladimir.balayan@feedzai.com} \\
    \And
    Pedro Saleiro \\
  Feedzai \\
  \texttt{pedro.saleiro@feedzai.com} \\
     \And
    Catarina Belém \\
  Feedzai \\
  \texttt{catarina.belem@feedzai.com} \\

        \And
     Ludwig Krippahl \\
   Universidade Nova FCT \\
   \texttt{a4338@fct.unl.pt } \\
  \And
    Pedro Bizarro \\
  Feedzai \\
  \texttt{pedro.bizarro@feedzai.com} \\

}
\begin{document}

\maketitle

\begin{abstract}
Machine Learning (ML) has been increasingly used to aid humans to make better and faster decisions.
However, non-technical humans-in-the-loop struggle to comprehend the rationale behind model predictions, hindering trust in algorithmic decision-making systems. Considerable research work on AI explainability attempts to win back trust in AI systems by developing explanation methods but there is still no major breakthrough. At the same time, popular explanation methods (\eg, LIME, and SHAP) produce explanations that are very hard to understand for non-data scientist persona.
To address this, we present JOEL, a neural network-based framework to jointly learn a decision-making task and associated explanations that convey domain knowledge.
JOEL is tailored to human-in-the-loop domain experts that lack deep technical ML knowledge, providing high-level insights about the model's predictions that very much resemble the experts' own reasoning. 
Moreover, we collect the domain feedback from a pool of certified experts and use it to ameliorate the model (human teaching), hence promoting seamless and better suited explanations.
Lastly, we resort to semantic mappings between legacy expert systems and domain taxonomies to automatically annotate a bootstrap training set, overcoming the absence of concept-based human annotations. 
We validate JOEL empirically on a real-world fraud detection dataset. We show that JOEL can generalize the explanations from the bootstrap dataset. Furthermore, obtained results indicate that human teaching can further improve the explanations prediction quality by approximately $13.57\%$.

\end{abstract}

\section{Introduction}
\label{sec:intro}

Notwithstanding the benefits of \ac{ML} commoditization in many areas, several state-of-the-art \ac{ML} methods convey recondite information about their predictions. The inability to comprehend these outcomes hinders trust in these systems. 

Research on \ac{XAI} focuses on improving the interpretability (or \textit{transparency}) of these systems.
Whilst the \ac{XAI} literature is booming with explainer methods \cite{Ribeiro2016, Lundberg2017, Plumb2018, Ribeiro2018, Zafar2019}, the large majority focus on the data scientist persona, creating low-level explanations that are too difficult to grasp to other non-technical personas
\cite{Ghorbani2019, Kim2018, melis2018towards}. 
This gap is witnessed in the recent impetus of feature-attribution-based explainers \cite{Ribeiro2016, Lundberg2017, Plumb2018}. These build explanations as lists of input features and associated feature scores\footnote{The feature's contribution to a given \ac{ML} prediction}. However, these can be too specific and too technical.
For this reason, explainer methods alike fail to address the information needs of \textit{domain experts}, which are deeply involved in many human-AI cooperative systems \cite{Ghorbani2019, melis2018towards}.

Instead, it is necessary to create explanations capable of conveying domain knowledge. This should reflect the semantic concepts related to the task of the human-in-the-loop, encoding information about the domain experts underlying reasoning upon decision-making. Although recent lines of research have begun exploring ``concept-based explanations'' \cite{Kim2018, Ghorbani2019, melis2018towards, panigutti2020doctor} these do not apply to high stakes decision-making tasks. 

\begin{figure}[hbtp]
    \centering
    \includegraphics[width=0.95\textwidth]{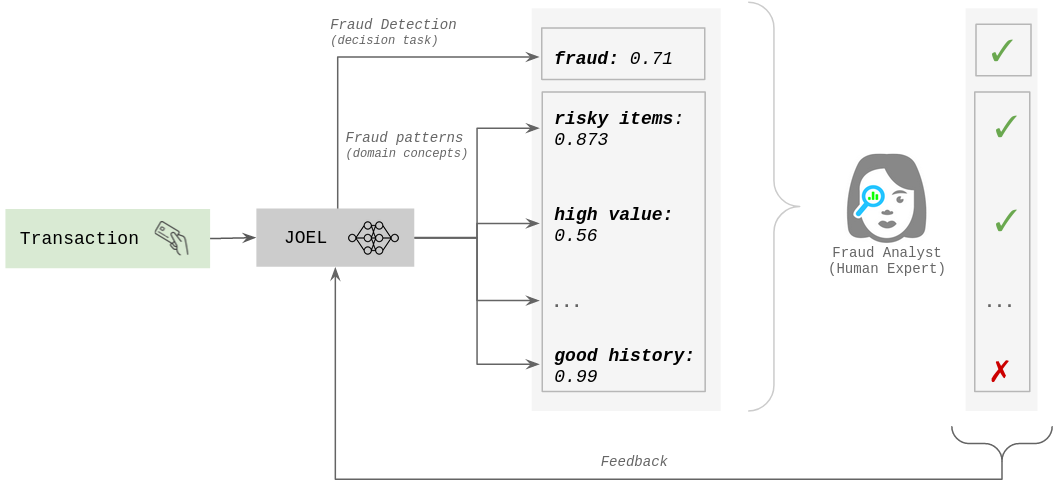}
     \caption{Example of JOEL's online loop of generating concept-based explanations and collecting human feedback in a fraud detection settings.}
    \label{fig:joel_human_teaching}
\end{figure}

In this work, we aim to strengthen human-AI decision-making systems by: (1) proposing a novel \ac{XAI} framework, JOEL, a self-explainable neural network that jointly learns semantic concepts and a decision associated with a predictive task \cite{melis2018towards, elton2020self}, and (2) exploiting the human-in-the-loop feedback about the decision task and explanations quality to continuously produce better suited explanations. Figure \ref{fig:joel_human_teaching} illustrates the proposed framework in a real world fraud detection setting.

The joint learning of the decision task and associated explanations leads to more robust and authentic\footnote{In this context, we refer to authenticity as being a true reason for a given \ac{ML} outcome.} explanations. 
We encode the \ac{ML} interpretability in an \ac{NN} model, hence allowing us to incorporate domain knowledge and to associate semantic concepts to the decision-making task \cite{Kim2018, Ghorbani2019, melis2018towards, panigutti2020doctor}. These high-level concepts may encode external information that while not present in the data itself may be very useful.

The human-AI proximity facilitates the creation of virtuous cycles where the model learns, through human (expert's) feedback, how to improve its explanations in a way that benefits the human at decision-making: the model learns the \textit{teacher}'s reasoning and with time it can foster faster and more effective decisions from the \textit{teacher}. 

Finally, we use a Distant Supervision approach to circumvent the low availability of human concept-labeled datasets\footnote{Manually creating these datasets carries abnormal costs and time efforts.\cite{melis2018towards}} when there are legacy expert systems in place (\eg, rule-based systems). This approach automates the labeling process based on semantic mappings between the existing legacy system and domain knowledge taxonomies. 

The summary of our contributions are as follows:

\begin{enumerate}
    \item We present a novel \textbf{self-explainable neural network} framework to mutually learn a decision task and an explainability task, consisting of associated high-level domain concepts that serve as explanations to the human-in-the-loop (section \ref{ssec:architecture}); 
    \item We exploit the \textbf{human-in-the-loop's expertise} to continuously tune the model on both tasks (section \ref{ssec:human_teaching});
    \item We propose and empirically validate the use of a \textbf{semi-supervised approach} for bootstrapping the proposed self-explainable neural network model when manually concept-based annotated datasets are unavailable (section \ref{ssec:distant_supervision}). 
    \item We show on a \textbf{real-world fraud dataset} that JOEL is able to jointly learn the semantic concepts and the fraud decision task. We observe an increase of the model's explainability performance of approximately $13.57\%$ after the incorporation of expert's feedback into the model (section \ref{sec:results}).
    
\end{enumerate}

\section{Methodology}
\label{sec:methodology}

In this section, we describe our framework for jointly learning a decision task and associated domain knowledge explanations. We assume the semantic concepts (used as explanations) will help the domain experts (end-users) reasoning throughout their decision-making process. Conversely, the framework benefits from the domain experts' feedback about which concepts justify their decisions. Hence, it can continuously improve both its predictive accuracy and explainability. We coin this framework \ac{JOEL}.

\subsection{Architecture}
\label{ssec:architecture}

We frame the problem of associating both semantic concepts and decision labels as one of finding an \textit{hypothesis} (learner), $h \in \mathcal{H}$, such that, for the same inputs, $x \in \X$, $h$ is able to simultaneously satisfy $h: \mathcal{X} \rightarrow \mathcal{S}$ and $h: \mathcal{X} \rightarrow \mathcal{Y}$ , where $\mathcal{S}$ is the set of semantic concepts and $\Y$ is the set of decisions (or classes) of the predictive task.

\begin{figure}[hbtp]
    \centering
    \includegraphics[width=\textwidth]{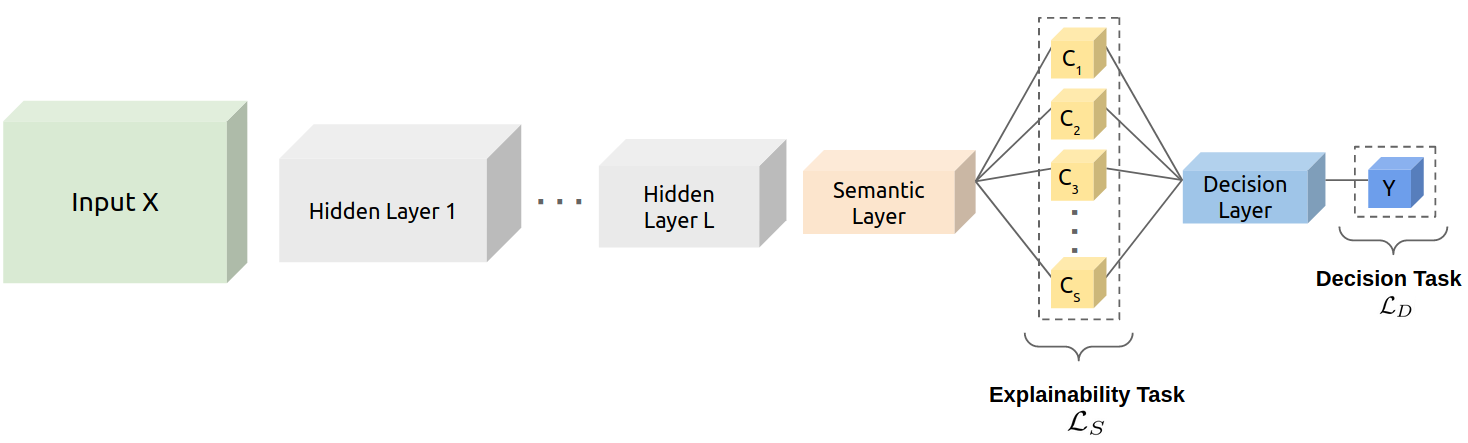}
     \caption{JOEL hierarchical architecture. The semantic layer (orange) produces the concepts, $C_s$ (yellow), which are the inputs for the decision layer (blue). The concepts (explanations of the predictive task) are also outputs of the network. Colors indicate layer type: input vector (green); hidden layer (grey); semantic layer (orange); decision layer (blue); output layers (white).}
    \label{fig:joel_hierarchical}
\end{figure}

Conceptually, \ac{JOEL} comprises three building blocks: (1) an \ac{NN}, (2) a semantic layer, and (3) a decision layer. \ac{NN}'s versatility facilitates the arrangement of these components in multiple ways. This work, however, delves into a hierarchical version (see figure \ref{fig:joel_hierarchical}): whose blocks are chained sequentially, \ie, the outputs of the \ac{NN} are fed as inputs to the semantic layer and its outputs are fed into the decision layer.
\ac{JOEL} outputs both decision predictions (last layer) and concept predictions. 

Vanilla \acp{NN} seek to model single-label tasks, such as fraud detection, or medical diagnosis. 
\ac{JOEL}'s hierarchical architecture can be seen as a multi-task materialization where the outputs of the $(L+1)^{th}$ layer are used both for a multi-label task (predict multiple semantic concepts) and a single-label task. 
By chaining the semantic and decision layers, we are able to encode external information about the domain which is not available in the feature data. 
This is particularly meaningful when the taxonomy of semantic concepts is intimately related to the decision task (\eg, a fraud taxonomy of fraudulent patterns is deeply correlated with the fraud detection task). 
Therefore, learning to accurately predict domain concepts is likely to also lead to better predictions and end-user decisions when it comes to the end-task.  

Similarly to other \ac{NN} architectures, our method learns through backward propagation of errors (\textit{backprop})\cite{Rumelhart1986LearningRB} 
and some variant of the gradient descent method. 
Unlike other \acp{NN} models, the joint learning approach in JOEL's method attempts to minimize both a \textit{decision loss}, $\Loss_D$, and a \textit{semantic loss}, $\Loss_S$. Mathematically, given  model's parameters, $\Theta = \left[\theta_1, \theta_2, ..., \theta_L, \theta_S, \theta_D\right]$, the outputs of the decision layer, $\delta_D(x, \Theta)$, and the outputs of semantic layer, $\delta_S(x, \Theta)$, the gradient with respect to the loss, $\Loss$, at the semantic layer is given by equation \ref{eq:multi_class_classification_contribution}.

\begin{equation}    
    \label{eq:multi_class_classification_contribution}
    \nabla_{\delta_S} \Loss(x, y, s) = \nabla_{\delta_S} \Loss_D(\delta_D(x, \Theta), y) + \nabla_{\delta_S} \Loss_S(\delta_S(x, \Theta), s)\\
\end{equation}

where: 

\begin{equation}    
    \label{eq:decision_mapping}
    \nabla_{\delta_S} \Loss_D(\delta_D(x, \Theta), y) = \dfrac{\partial\delta_D(x,\Theta)}{\partial \delta_S}.\nabla_{\delta_D} \Loss_D(\delta_D(x, \Theta), y))
\end{equation}

The decision of which loss functions to use within each task depends on the nature of the task. Since the semantic layer corresponds to a multi-labeling task, we use a \textit{sigmoid} function for each individual entry of the output, before using it in the loss function. 
To find the mapping that simultaneously satisfies $h: \mathcal{X} \rightarrow \mathcal{Y}$ and $h: \mathcal{X} \rightarrow \mathcal{S}$ for a given input vector, $x \in \X$, we mutually minimize the (categorical) cross-entropy for both predictive tasks. For an input vector, $x \in \X$, a set of domain concepts, $s \in \mathcal{S}$, and decision labels, $y\in\Y$, we have that:

\begin{align}
\label{eq:loss_functions}
    \Loss_D(x,y) &= -\sum_{i=1}^{|\Y|} y_i \log \left[ \text{softmax}(\delta_{D}(x, \Theta)_i)\right]\\
    \Loss_S(x,s) &= -\sum_{i=1}^{|\mathcal{S}|} s_i \log \left[ \text{sigmoid}(\delta_{S}(x, \Theta)_i) \right]
\end{align}

The real ``intepretability'' of concept-based explanations is debatable \cite{melis2018towards}, mirroring the incertitude within the \ac{XAI} field. From a fundamental standpoint, using semantic concepts to explain decisions does not necessarily signify the \ac{NN} leverages their meaning in its predictive task. 
However, by design, we do expect the concepts to be intimately interrelated with the decision task. In this view, JOEL's explanations may actually bestow promising guidance to a skillful persona with no ML technical knowledge, thereby improving its efficiency and efficacy.

\subsection{Human Teaching}
\label{ssec:human_teaching}

Human teaching closes the human-AI feedback loop in algorithmic decision-making systems. It is often the case that interactions within these systems are uni-directional: a human (typically, a domain expert) interacts with an \ac{ML} system to make a decision. However, while the \ac{ML} system directly influences human decisions, the reverse is not verified, as most systems do not promptly adapt to human behavior. 
Instead, most systems are offline and it is only after a certain period of time that a new model is trained and adapted to the collected knowledge.

\ac{JOEL} tackles these limitations and tailors the explanations' quality to be on par with the domain expert's need. Since the explanations are meant to help the domain experts to better perform their task, these should also resemble the experts' own reasoning. As a result, the integration of expert feedback into the model's learning process becomes vital to ensure that the explanations remain interpretable and that keep fostering more efficient and effective decisions.

Figure \ref{fig:joel_human_teaching} shows this mechanism in an illustrative example of a fraud detection task, where the main goal is to classify financial transactions as being fraudulent or not. After receiving a transaction, \ac{JOEL} infers predictive scores for both the fraud decision and the fraud patterns (semantic concepts). Using this information (explanation), a fraud analyst (domain expert) reviews the transaction and indicates which fraud patterns she based her decision. Lastly, \ac{JOEL}'s learnable parameters get updated based on the expert's feedback.

In many real-world settings, human expertise aims to disambiguate inputs for which the model is very uncertain. In these cases, JOEL can exploit this short-term feedback to improve human-AI system performance. Our method's hierarchical structure 
is likely to encode extra information based on this feedback. Therefore, rapidly improving both its predictive accuracy and the explanations quality.

After bootstrapping JOEL using an initial concept-based annotated dataset, JOEL enters a \textit{human teaching stage}, interacting with humans-in-the-loop: providing explanations and collecting human feedback about the generated explanations. Every $b$ feedback instances, JOEL uses them to adjust its parameters through \textit{backprop}. To prevent eventual model deterioration (\eg lower human aptitude for the task), we incorporate a quality control precondition for running this stage: to adhere to the pool of human teachers in the human-teaching stage, each human must meet a minimum accuracy-level (or other relevant qualitative measure).


\subsection{Distant Supervision}
\label{ssec:distant_supervision}

Training neural networks requires large amounts of data in order to attain reasonably good performance \cite{roh2019survey}. Multi-label methods further exacerbate this data necessity. 
On top of that, difficulties associated with the collection and creation of concept-annotated datasets make \ac{NN} explainers based on semantic concepts infeasible in many practical settings. 

JOEL demands large datasets to be able to generalize well. When there is no data available (or the data is insufficient) and information based on legacy expert systems is available, we propose to bootstrap the model using a semi-supervised learning approach known as \textit{Distant Supervision} \cite{mintz2009distant, go2009twitter, zeng2015distant}.
Often times, these expert systems enclose semantic knowledge that can be associated with the concepts from the domain taxonomy. Therefore making it possible to automatically derive concepts for each datapoint without the need for human labels. 
While it still requires human effort to create the legacy systems-taxonomy mappings, the effort is negligible when compared with the effort of manually annotating a large dataset.

As an example consider the fraud prevention domain involving a rule legacy system. Using the similarities between the domain knowledge conveyed in the rules, we create a \textit{rule-concept mapping} to automatically associate rules to concepts in a fraud taxonomy. Table \ref{reasons_rules_mapping} shows examples of such mappings. After proper validation of the mapping by a fraud expert, we use it to automatically label payment transactions in bulk. Specifically, consider a payment transaction $X$ for which the legacy system triggered the two rules in the table. The first rule maps to the ``Suspicious Items'' and the second rule maps to ``Suspicious Customer'' and ``Suspicious Payment''. 
Thus, the Distant Supervision technique automatically annotates transaction $X$ with ``Suspicious Items'', ``Suspicious Customer'', and ``Suspicious Payment''.

\begin{table}[htpb]
    \caption{Example of two \textit{rule-concept mappings} in a fraud detection setting.}
    \setlength\extrarowheight{2pt}
    \begin{tabularx}{\textwidth}{XX}
        \hline
        \textbf{Rule Description} & \textbf{Mapped Concepts} \\
        \midrule
            Order contains risky product styles. & Suspicious Items\\
            User tried \textit{n} different cards last week. & Suspicious Customer, Suspicious Payment\\
        \hline
    \end{tabularx}
    \label{reasons_rules_mapping}
\end{table}
Although semi-supervised techniques improve scalability, one still requires a human-labeled dataset to empirically validate the results. As we will see in the ensuing section, we conduct an empirical validation of the proposed method using Distant Supervision to automatically label a dataset with $9.3$ million instances, followed by a validation in a smaller human-labeled dataset with $1561$ transactions.

\section{Experimental Setup}
\label{sec:exp_setup}

We validate our explanation method empirically in a real Human-AI fraud prevention system. This system comprises a rule legacy system, a deployed \ac{ML} model, and a pool of fraud analysts. In this context, when a payment transaction first arrives, it undergoes a rule system, possibly triggering pre-defined business rules (\eg, payment amount is anomalous or the IP of an online payment transaction is blacklisted). Subsequently, an \ac{ML} model assigns the transaction a risk score. The prevention system immediately issues a decision for the most extreme scores (accept low-risk and reject riskier ones). Accordingly, transactions whose score is within a business-defined review band require human intervention. In that case, fraud analysts inspect the transaction data
, fraud risk, and the explanations, to decide whether to accept or reject the transaction. 

We evaluate JOEL in a two-step approach. We first run a grid search over a ``noisy dataset`` to determine the best architectural and learning decisions (\eg, number of hidden layers, number of neurons at each layer). In total, we train $71$ models. Then, we fix the decisions for the best model and we run another grid search over the ``ground truth dataset`` (based on human labels) to determine the best configuration possible for a human teaching setup and whether the models benefit human teaching. Finally, throughout these experiments, we examine the different trade-offs emerging from jointly learning two distinctive tasks: fraud detection (single-label task) and creation of explanations based on fraud patterns (multi-label task). We coin the first stage \textit{Distant Supervision} and the second \textit{Human Teaching}.

\textbf{Dataset}: We use a privately held online retailer fraud detection dataset, comprising approximately $9.3$ millions of payment transactions of which only $2\%$ represent fraudulent behaviors. For our experiment, we divide it into four sequential subsets: train, validation, test, and production. These represent $47\%$, $16\%$, $9\%$, $28\%$ of the dataset and exhibit $2.5\%$, $1.7\%$, $1.5\%$, $1.5\%$ of fraud rate (prevalence), respectively. While training JOEL, we use the validation set to apply early stopping and set the outputs' thresholds. Then, we perform model selection in the test set, after which only the best model is evaluated in a production set. The production and test sets are apart by two months, thus mimicking practical settings where a large time gap exists between model selection and deployment. 

\textbf{Domain Knowledge taxonomy}: With the help of skilful fraud analysts, we created a fraud taxonomy of suspicious patterns to guarantee that the provided explanations closely resemble their reasoning process throughout transactions reviewing. Rather than creating prosaic explanations, the taxonomy consists of keywords that correspond to certain financial transaction patterns such as ``malformed addresses'' or ``malformed name``. Furthermore, the taxonomy is hierarchic, grouping patterns alike into more general concepts, for instance, by grouping the previous two patterns into the single broader concept ``suspicious billing shipping''.
Semantically, these concepts refer to patterns involving specific information about the transaction and aim at pinpointing the most relevant information (\eg, evidence of wary behaviors) that the analysts look for when reviewing. As an example consider the concept ``suspicious billing shipping'' which aims to guide analysts' attention to the information related with shipping or/and billing information and look for dubious aspects (\eg mismatch between addresses, and malformed address). 

\textbf{Distant Supervision Dataset} (or machine-labeled dataset): In the absence of a human-labeled dataset, we use the Distant Supervision approach to automatically annotate  the initial single-label dataset (exclusively annotated with fraud labels). We extract the already existing information of a legacy rule system (that encodes high-level domain information) and map it into the fraudulent patterns in the taxonomy. We had domain specialists supervised this mappings, which totals $300$ associations.
The result is a multi-label dataset: where each row is jointly associated with the fraud label (decision task) and fraud patterns (semantic concepts). Henceforth, given that these annotations are proxies of the true associated concepts, we dub them ``noisy labels''. We observe a poor rule coverage of non-fraudulent patterns (\eg ``Nothing Suspicious'', ``Good Customer History'', ``All Details Match''), which emphasizes the limitation of this approach on both the available rules.  


\textbf{Human Labeled Dataset}: A robust validation of our method should not depend exclusively on machine-labeled datasets (``noisy datasets''), for they are not guaranteed to reflect the real data. Thus, we ran a human-labeling campaign involving three in-house fraud experts, where we mimic their original reviewing process. After each transaction review, the system prompts the analysts about the fraud patterns (or legitimate patterns) perceived during the transaction appraisal. Analysts select the concepts from the pool of concepts determined in the fraud taxonomy, and also the fraud decision. This human effort yielded $1561$ manually labeled transactions. Henceforth, we dub these labels as ``ground truth'' labels. 

\textbf{Evaluation Metrics}: JOEL comprises two distinct tasks: a single-label task focused on discerning fraudulent from non-fraudulent transactions, and a multi-label task focused on producing a complete and informative explanation for the former. Complying with the existing business constraints, which aim to reduce the probability of negatively affecting legitimate users, we set out to evaluate the first task in terms of Recall at $3\%$ \ac{FPR}.
The evaluation results (after training and validation) require an aggregate measure of JOEL's generalization at the multi-label task in holdout datasets. We use the mean \ac{AUC} over all predicted concepts because \ac{AUC} gives a general idea about the overall performance for a single label. 

\textbf{Baseline}: We compare our hierarchical structure with the vanilla multi-label \ac{NN}. In this baseline, instead of considering a hierarchical structure, we set both the decision and the semantic layer at the same level.

\textbf{Hyperparameter Optimization best model criterion}: From all the evaluated hyperparameter configurations (both for baseline and hierarchical structure), we select the one with highest performance at the fraud detection task. This decision guarantees business constraints are met regardless of the explanations. In case of ties, we discern the best model based on the performance at the second task - explanations accuracy, measured on the Distant Supervision dataset.
 
\textbf{Human Teaching}: The second experiment uses the human-teaching stage, the human-labeled dataset, and the best architectural decisions as determined in the first model selection stage. We carry out a small experiment to obtain proxy measures of the fraud analysts' accuracy (prior to being shown explanations). All three analysts satisfied the minimum accuracy level of $0.5$ set for this experiment (with $0.636$, $0.612$, and $0.597$ accuracy levels), and were, therefore, integrated in the teachers pool of the human-teaching stage. Afterwards, we split the human-labeled dataset into training, validation, and test sets. The first $800$ transactions are used for training and the ensuing $200$ for early stopping, threshold calibration, and model selection. Then, we re-train the best performing model using both the training and validation sets. We measure its generalization performance in an holdout test set composed of $561$ transactions.

\section{Results and Discussion}
\label{sec:results}

\begin{figure}[h]
    \centering
    \includegraphics[width=\textwidth]{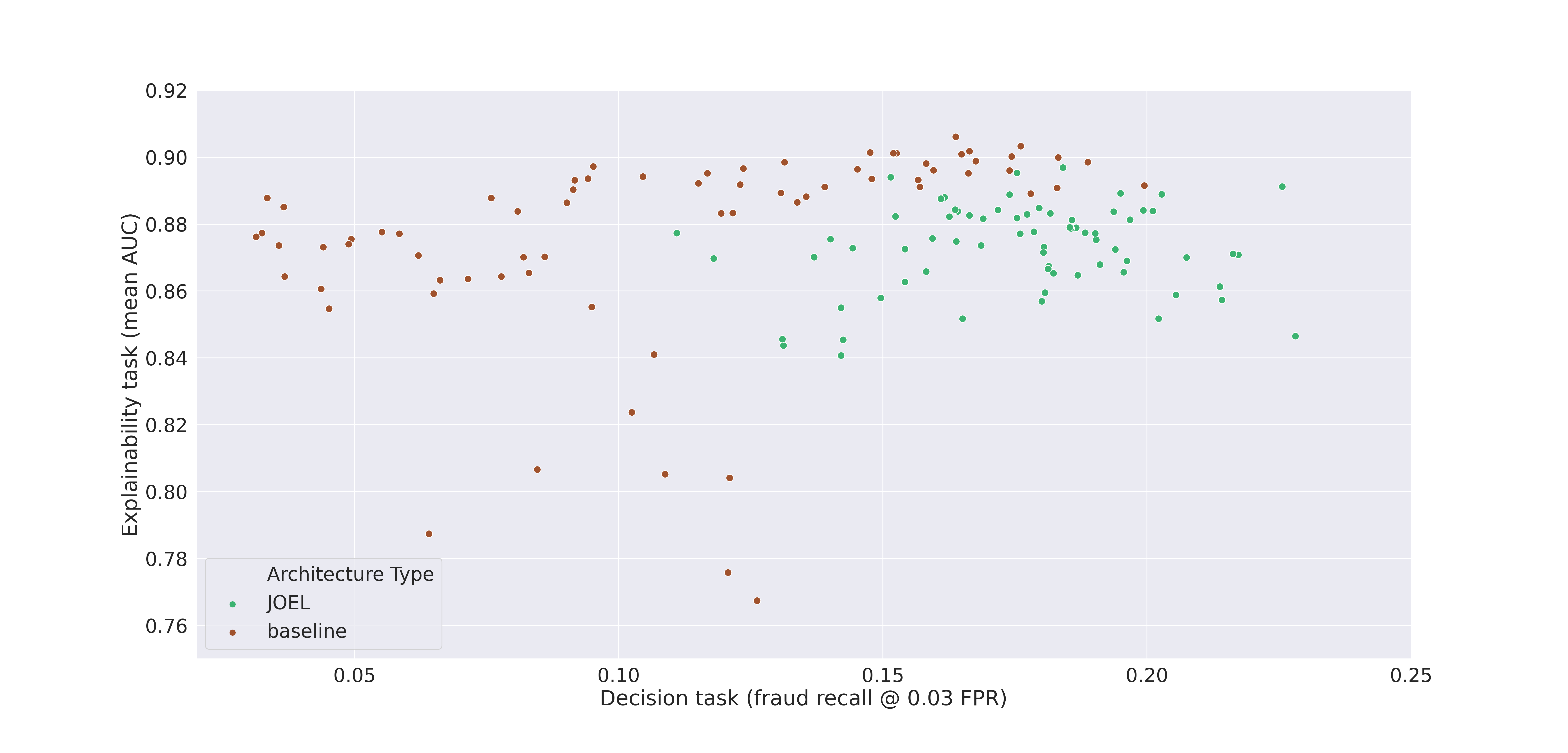}
     \caption{Trade-off between the decision task (fraud detection) and explainability task (concepts prediction) discriminated by architecture type.}
    \label{fig:expl_decision_tradeoff}
\end{figure}

Figure \ref{fig:expl_decision_tradeoff} shows the obtained trade-offs between decision task (fraud detection) and explainability task (predicting associated fraud concepts) for both JOEL and the baseline. In total, we trained $71$ distinct models for each architecture. 
In terms of the decision task, JOEL-derived architectures seem to exhibit a propensity for achieving high results in the predictive task, while also achieving competitive mean \ac{AUC} values for the explainability task. Conversely, although baseline models exhibit in some cases higher performance in the explainability task they fall short at the fraud prediction task. In fact, only one baseline model ranks in the the top $15$ best performing models on the decision task (ranks $12^{th}$ with $19.95\%$ of fraud Recall at $3\%$ \ac{FPR}).

Thus, given the business constraints, we select the best model in terms of fraud recall at $3\%$ \ac{FPR}, achieving $22.81\%$ Recall on the test set. The selected JOEL model has a total of $5$ hidden layers (with $128$, $128$, $64$, $64$, and $32$ neurons), a semantic layer (with $32$ neurons), and a decision layer (with $14$ neurons). At each hidden layer, we use the \textit{ReLU} activation function \cite{zhang2018multiple}, followed by a dropout mechanism \cite{srivastava2014dropout} (with rates of $40\%$, $40\%$, $30\%$, $30\%$, $20\%$). We also apply a dropout rate of $20\%$ to the semantic layer). Finally, we use the out-of-the box \textit{Adam} optimizer \cite{kingma2014adam} and set the learning rate to $0.0001$.

\textbf{Distant Supervision results:} We use Distant Supervision to bootstrap our models with ``noisy labels''. As a result, the quality of the model's predictions is directly correlated with the rule-concepts mapping: if a concept does not have a matching rule, then the model will suffer from the cold-start problem\footnote{Other semi-supervised techniques discussed in section \ref{sec:related_work} can be used to overcome the cold-start problem.}. Even if there exist rules that map to each concept, if they exhibit small prevalence in the dataset, then it is likely that the model struggles to learn them. This concepts' misrepresentation reflected on the obtained results. Table \ref{tab:model_performance_distant_supervision} shows the results of the best JOEL model (with $0.8446$ mean \ac{AUC}) on the hold-out Distant Supervision production dataset discriminated by semantic concept. Although the model attains \ac{AUC}'s above $0.80$ for most concepts, for two of them the performance is worse than $0.65$. 
We hypothesize this to be related with the concept's misrepresentation. In particular, we emphasize that the rule system is tailored for anomalous behavior detection, with only a small fraction of the rules mapping to legitimate behaviors. Therefore, with the exception of the ``Suspicious Delivery'' concept, the model exhibits fairly high \ac{AUC} values for fraud concepts. Since only $5$ rules map to ``Suspicious Delivery'', all of which underrepresented, the model is not able to generalize it.

\begin{table}[btph]
\centering
\caption{Model performance on production dataset, trained with Distant Supervision.}
\begin{tabular}[t]{lc}
\toprule
\textbf{Concept} & \ac{AUC} \\
\midrule
Suspicious Items            &   0.8499  \\
Suspicious IP               &  0.9190   \\
Suspicious billing shipping &   0.8921  \\
Suspicious Delivery         &    0.6467 \\
Suspicious Email            &   0.9376   \\
Suspicious Payment          &   0.9156   \\
\bottomrule
\end{tabular}
\quad
\begin{tabular}[t]{lc}
\toprule
\textbf{Concept} & \ac{AUC}  \\
\midrule
Suspicious Device           &    0.8862 \\
High speed ordering         &   0.9468  \\
Suspicious Customer         &   0.9401  \\
All details match           &  0.6470 \\
Nothing suspicious          &   0.7169 \\
Good customer history       &  0.8370  \\
\bottomrule
\end{tabular}
\label{tab:model_performance_distant_supervision}
\end{table}


\textbf{Human teaching results}: In the second stage of our experiment, we start with the evaluation of the best model in the previous stage on a human labeled test set, containing $561$ instances. This evaluation comprises no hyperparameter or structural tweaks. We then compare this approach with one where the model tweaks its hyperparameters based on the human teachers' feedback. For this teaching stage, we increase the model's learning rate from $0.0001$ to $0.05$, hence causing the model's weights to be updated more aggressively. Table  \ref{tab:model_performance_human_teaching}) shows the obtained \ac{AUC} results for the former untweaked model (``no human teaching'') and also for the model after the human teaching stage (``human teaching''). We observe a small improvement of approximately $0.04$ for the ``Suspicious Delivery'' and ``All details match'' concepts, for which the Distant Supervision approach had an initially worse performance. 
Interestingly, it is also possible to observe a detriment in the predictive performance of the concept ``High speed ordering'', ``Suspicious Customer'', ``Nothing Suspicious'', and ``Good customer history''. One possible cause of this result may be incongruent mental representations between human experts. A possible way to mitigate it, would be to implement an agreement-based filtering technique that would be responsible for filtering out the human feedback based on the agreement of multiple teachers. 

\begin{table}[htbp]
\centering
 \caption{Performance of the best bootstrapped model on human-labeled test set with and without human teaching.}
\begin{tabular}[t]{lcc}
\toprule
{Concept} & No Human Teaching (AUC) & Human Teaching (AUC) \\
\midrule
Suspicious Items            &   0.2911    & 0.7119 \\
Suspicious IP               &   0.4760   &   0.7215\\
Suspicious billing shipping &   0.5943   &   0.7531\\
Suspicious Delivery         &  0.5962    &  0.6368 \\
Suspicious Email            &  0.6231    &   0.7222\\
Suspicious Payment          &   0.6971   &  0.7170 \\
Suspicious Device           &   0.7348   &   0.7799\\
High speed ordering         &  0.7365    &  0.6874\\
Suspicious Customer         &   0.4621   &  0.3640 \\
All details match           &    0.3496   & 0.3976 \\
Nothing suspicious          &   0.4151   & 0.3425  \\
Good customer history       &  0.4250  &   0.3266  \\
\midrule
Mean & 0.5393 &  0.6125\\
\bottomrule
 \end{tabular}
\label{tab:model_performance_human_teaching}
\end{table}




Overall, we can observe that (1) the pre-trained model with Distant Supervision approach is able to generalize the explanations, and (2) the human feedback enhances, on average, the model's explanation quality, improving the predictive performance of concepts it initially struggled to predict.


\section{Related Work}
\label{sec:related_work}

The use of human-friendly semantic concepts is becoming more popular, specifically, in the computer vision domain. For instance, Kim \etal \cite{Kim2018} proposed Testing with Concept Activation Vectors (TCAV), a concept-based explainer that weighs the importance of a user-defined concept to a classification result (\eg, how sensitive the prediction of ``bird'' is to the presence of concepts like ``feathers'' or ``beak''). 
Recently, this method has also been applied to the natural language domain, for example, to measure how neural network activation units react to different human values-laden terms such as ``respective'', ``abusive'', ``straight'' among others  \cite{Hutchinson2019}.  
Also in the computer vision domain, Ghorbani \etal \cite{Ghorbani2019} propose Automated Concept-Based Explanation (ACE), that  extracts concepts directly from the input data without any external source. It first obtains different embeddings for different image segments and then applies a clustering algorithm in the embeddings space to identify the concepts. 

In another perspective, Melis and Jaakkola \cite{melis2018towards} include the interpretability straight into the \ac{ML} model architecture (dubbed self-explainer concept-based model). They use autoencoders to generate the basis concepts and their relevance scores from low-level input features, which then serve as input to the predictive task.
In another vein, \textit{DoctorXAI} \cite{panigutti2020doctor} leverages domain knowledge to produce more comprehensive explanations in medical diagnosis tasks. This domain information is used to guide the explanation generation process of a low-level explainer, LIME \cite{Ribeiro2016}.

All in all, most concept-based methods in the literature are specific to computer vision. Indeed, there is currently no research work in what it comes to high-level concept-based explanations for decision-making tasks. While \textit{DoctorXAI} \cite{panigutti2020doctor} uses concepts to create explanations for a decision-making setting, the explanation itself does not convey any domain knowledge about the decision but rather about the feature space. Hence, it fails to address information needs of non-data scientist humans-in-the-loop.

A crucial aspect in interpretable concept learning concerns the data quality, since many of the previous methods demand for largely annotated datasets. Unfortunately, obtaining enough human annotations can be difficult in practice due to the time and efforts required.  
In this regard, semi-supervised learning approaches constitute a promising bypass to the label shortage \cite{mintz2009distant, go2009twitter, ratner2016data, srivastava-etal-2017-joint, Hancock2018nlpexplanations}. Most of these approaches foster the programmatic labeling of datasets. One such example is the Distant supervision technique \cite{mintz2009distant, go2009twitter}, which leverages existing knowledge bases (\eg, rule systems) and, through the heuristically creation of semantic mappings between the knowledge bases and the concepts, facilitates the automatic (weak) annotation of the dataset. Another emerging technique in the natural language domain is the joint use of written explanations and labeling functions to train a generative labelling model \cite{srivastava-etal-2017-joint, Hancock2018nlpexplanations}. In this context, a small subset is annotated with natural language explanations and decision labels. These, along with the instances, are then used to hot-start and parametrize the labeling functions that will be responsible for generating the labels. In our experiments, we use the distant supervision approach to exploit the already available vestiges of a legacy rule-based system in the data. We leave the exploration of other approaches for future work. 

\section{Conclusion and Future Work}
\label{sec:conclusion}

In this work, we propose JOEL, an \ac{NN}-based framework for jointly learning a decision task and associated domain knowledge explanations. Simultaneously, our framework exploits the decision-makers feedback regarding not only their decision but also the concepts that justify their decision. Initial experiments show that using human teaching can lead to significant gains in predictive accuracy and explainability. Moreover, we show that JOEL is able to overcome the cold-start problem and generalize from a machine-labeled dataset when using a Distant Supervision technique. 
Future directions include validating JOEL in other settings (\eg, in the healthcare domain), robustly assessing the framework's true impact and utility for the users, exploring distinct \ac{NN} variants (\eg, architectures, loss functions), and exploring incremental learning to continuously adapt to new knowledge.

\section{Acknowledgements}
\label{sec:achnowledgements}

The project CAMELOT (reference POCI-01-0247-FEDER-045915) leading to this work is co-financed by the ERDF - European Regional Development Fund through the Operational Program for Competitiveness and Internationalisation - COMPETE 2020, the North Portugal Regional Operational Program - NORTE 2020 and by the Portuguese Foundation for Science and Technology - FCT under the CMU Portugal international partnership.

\small
\bibliographystyle{unsrt}
\bibliography{main}

\end{document}